\documentclass[10pt,twocolumn,letterpaper]{article} 

\usepackage{avss}
\usepackage{times}
\usepackage{epsfig}
\usepackage{graphicx}
\usepackage[table,xcdraw]{xcolor}
\usepackage{amsmath}
\usepackage{amssymb}
\usepackage{threeparttable}
\usepackage{tabularx}
\usepackage{subfigure}
\usepackage{hyperref}
\hypersetup{
    colorlinks=true,
    linkcolor=red, 
    citecolor=blue, 
    urlcolor=magenta    
}

\def\BibTeX{{\rm B\kern-.05em{\sc i\kern-.025em b}\kern-.08em
    T\kern-.1667em\lower.7ex\hbox{E}\kern-.125emX}}

\avssfinalcopy 

\ifavssfinal\fi
\begin{document}
\title{

Enhancing Vehicle Re-identification and Matching for Weaving Analysis
}

\author{Mei Qiu , Wei Lin, Stanley Chien, Lauren Christopher, Yaobin Chen, and Shu Hu\\
Purdue University Indianapolis\\
723 West Michigan Street, SL-160, Indianapolis, Indiana 46202, USA\\
        {\tt\small \
        meiqiu@iu.edu,\{schien,lauchris,ychen\}@iupui.edu,\
        hu968@purdue.edu}}


\maketitle
\thispagestyle{empty}
\let\thefootnote\relax\footnote{979-8-3503-7428-5/24/\$31.00 ©2024 IEEE}


\begin{abstract}
Vehicle weaving on highways contributes to traffic congestion, raises safety issues, and underscores the need for sophisticated traffic management systems. Current tools are inadequate in offering precise and comprehensive data on lane-specific weaving patterns. This paper introduces an innovative method for collecting non-overlapping video data in weaving zones, enabling the generation of quantitative insights into lane-specific weaving behaviors. Our experimental results confirm the efficacy of this approach, delivering critical data that can assist transportation authorities in enhancing traffic control and roadway infrastructure.
More information about our dataset can be found here: \href{https://github.com/qiumei1101/VeID-Weaving.git}{\color{blue}VeID-Weaving}.
\end{abstract}

\section{Introduction}
Surveillance cameras, strategically positioned along highways by state transportation departments, are crucial for monitoring traffic conditions. With their adjustable viewing angles and zoom capabilities, these cameras can focus remotely on specific highway segments, making them valuable for analyzing weaving patterns. However, they also present challenges for highway management organizations, as discussed in prior studies \cite{golob2004safety, shoraka2010review}. Highway weaving analysis is vital for traffic management and road improvement initiatives. It involves assessing vehicle lane changes between an entry and the subsequent exit ramp. This analysis calculates the percentages of vehicles that enter, exit, or continue past the exit ramp, utilizing short-term data collection at specific intervals. This process helps identify weaving patterns—lane changes within this section, under the assumption that there are no mid-section entries or exits. Such patterns are known to contribute to congestion and accidents during peak traffic periods.

\begin{figure}[t!]
  \centering
  \includegraphics[scale=0.24]{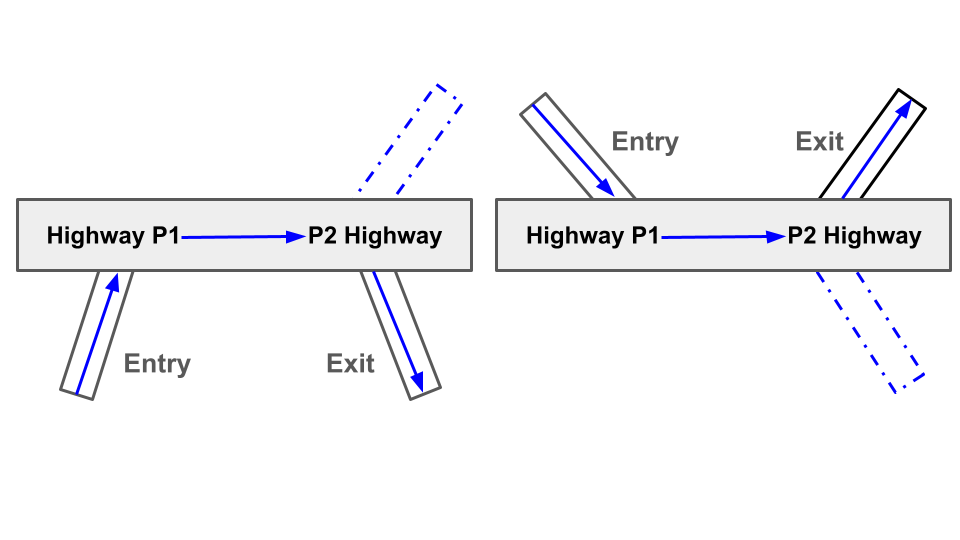}
  \caption{\textit{Weaving examples. Vehicles at P1 can either come from the ramp or the highway. As vehicles move to P2, they can stay on the highway or exit to the ramp}.}
  \label{fig:weaving_examp}
\end{figure}

For enhanced management, authorities require a quantitative understanding of these weaving patterns at the lane level. This information is critical for making informed decisions regarding route guidance, lane additions, or modifications in road geometry. Figure \ref{fig:weaving_examp} illustrates two typical weaving areas: where the highway merges with an entry ramp at P1 and where it splits with an exit ramp at P2. By correlating the vehicles entering at P1 with those exiting at P2, specific weaving patterns can be discerned. Given the common issue of non-overlapping camera coverage at P1 and P2 and the challenge of capturing clear images of distant vehicles, effective analysis demands innovative solutions to address the issues of non-overlapping fields of vision and small-vehicle detection.

Traditional traffic weaving data collection methods, like manual counting and video analysis at points P1 and P2 or through license plate detection, are laborious, error-prone, and expensive. Cellphone location data from third parties tracks only a fraction of vehicles with limited accuracy, making it unsuitable for detailed weaving analysis. Marczak et al. \cite{marczak2016empirical} pioneered research on lane changing behavior, sparking extensive investigation into weaving sections. Our study distinguishes between zone-level and lane-level methodologies. Zone-level analysis segments highways into zones for monitoring vehicle movements, requiring advanced algorithms and precise data \cite{ouyang2022traffic, xu2020modeling}. Lane-level analysis provides detailed insights into specific weaving patterns but may overlook interactions across lanes and demands accurate lane detection algorithms \cite{ouyang2023effects, arman2024empirical}.

Challenges in highway weaving studies include difficulties in data collection, vehicle tracking, matching accuracy, result generalizability, computational requirements, and the impact of video resolution quality. These challenges are compounded by data collection limitations, such as suboptimal camera angles and physical obstructions, the need for robust tracking algorithms, precise matching, limited applicability of findings, computational resources, and video quality's influence on analysis.

Our study introduces an innovative approach to estimate lane-level weaving by using two strategically placed cameras at points P1 and P2, which simultaneously monitor traffic flow. This method involves counting vehicles within predefined Regions of Interest (ROIs) on a lane-by-lane basis, achieving over 90\% accuracy in vehicle detection and counting. However, a major challenge arises in matching vehicles between non-overlapping video feeds from cameras positioned at a distance from one another. Typically, only a subset of vehicles is accurately matched based on their identifiable features. We consider these matched vehicles as a representative sample of the total vehicles detected, and accordingly, we present our weaving analysis results, acknowledging a certain margin of error.
The contributions of our work can be summarized as follows:
\begin{itemize}
\item We have developed a sophisticated framework to enhance our current system for analyzing lane-level vehicle behavior in weaving areas, using trajectory data from non-overlapping cameras.
\item We have created a large vehicle Re-Identification (ReID) dataset from 9 weaving areas using 16 highway cameras and 2 drone cameras. \textbf{Our dataset, with 4,902 unique vehicles and 78,979 images, is the first of its kind for highway weaving areas and offers more diversity in image sizes and resolutions, posing new challenges for vehicle ReID research.}
\item We analyzed lane-level traffic in five weaving areas using video data from various times of day: morning, noon, and later afternoon, finding that drivers' routes and time significantly impact lane-change frequency and locations. 
\item
Our analysis used deep and spatial-temporal feature matching, with a customized ReID model for feature extraction and an adapted the Hungarian algorithm \cite{kuhn1955hungarian} for vehicle matching.
\end{itemize}

\begin{figure*}
  \centering
  \includegraphics[scale=0.5]{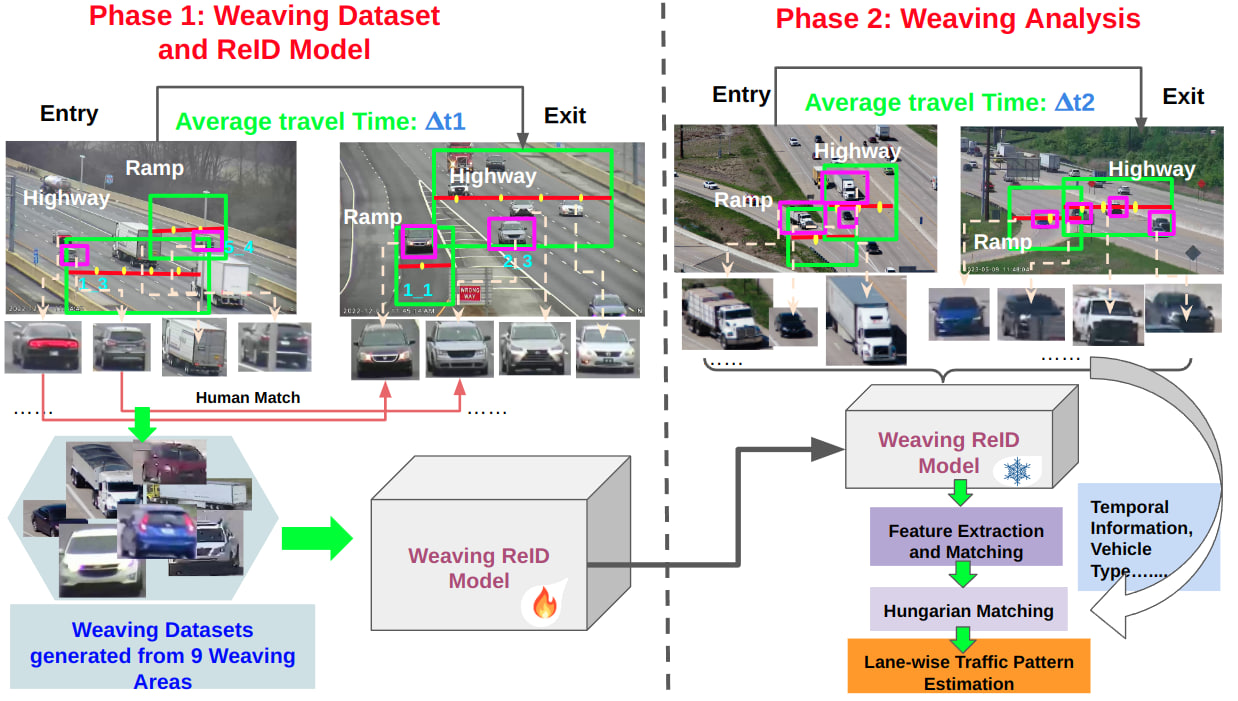}
  \caption{\textit{The framework of weaving analysis  consists of \textbf{two phases:Weaving Dataset and Weaving ReID Model and Weaving Analysis}. (Left) Phase 1: We create a custom weaving ReID dataset by matching vehicles across nine specific weaving areas. Subsequently, we train our custom weaving ReID model on this dataset. This phase focuses on establishing the groundwork for vehicle identification and tracking.
(Right) Phase 2: 
We match vehicles across weaving areas using the ReID model, extracting distinctive features from vehicle images. These features, combined with spatial-temporal information, are used in our Hungarian Matching module to derive lane-specific weaving patterns for comprehensive analysis. More details are explained in Section \ref{sec:method1} and \ref{sec:method2}.
}}
\label{fig:flow}
\end{figure*}
\section{Background}
\subsection{Lane-wise vehicle tracking in optimal ROIs in each weaving area}
In their study \cite{Mei2024}, they introduced a real-time system designed for lane-specific vehicle counting utilizing a single camera. This system adapts to various unfixed camera positions on highways by initially learning a comprehensive set of parameters, including road boundaries, lane centers, traffic flow direction, and lane margins, tailored to the current camera view. Should the camera's viewpoint change, the system automatically resets and adapts to the new perspective by learning a fresh set of parameters.

To overcome the challenge of inconsistent vehicle detection accuracy across different areas of the input image, the system identifies and learns from multiple strategically chosen Regions of Interest (ROIs). These ROIs are selected for their high potential in providing accurate vehicle detection, significantly boosting the overall performance of the vehicle tracking and counting system.

\subsection{Related Works}
\smallskip
\noindent
\textbf{Vehicle Detection in Surveillance Videos.}
Vehicle detection in surveillance videos plays a pivotal role in traffic monitoring, security operations, and urban planning. Recent advances in deep learning, especially through convolutional neural networks (CNNs), have significantly improved the accuracy and efficiency of such detections. Widely adopted models, including Faster R-CNN \cite{ren2015faster}, YOLO (You Only Look Once) \cite{redmon2016you}, and SSD (Single Shot Multi-box Detector) \cite{liu2016ssd}, excel at learning complex features directly from the data. Nonetheless, these models still face challenges in ensuring reliable detection in adverse conditions, handling occlusions, managing cluttered scenes, and scaling effectively across extensive surveillance networks.

\smallskip
\noindent
\textbf{Cross-Camera Vehicle Tracking.}
Cross-camera vehicle tracking in surveillance systems presents a complex challenge, necessitating the tracking of vehicles across non-overlapping camera views. This is typically addressed using multi-object tracking (MOT) frameworks, which integrate detection, data association, and trajectory optimization \cite{wu2021multi, yang2023cooperative}. These frameworks work by linking vehicle detections from separate cameras into unified tracks across multiple views. Robust appearance modeling and vehicle re-identification are crucial for ensuring the consistency of these tracks. Ongoing research focuses on creating robust and efficient methods to track vehicles across varied camera views in intricate settings \cite{li2023adaptive, gloudemans2024so}.

\smallskip
\noindent
\textbf{Vehicle Re-identification in Surveillance Videos and Datasets.}
Recent advancements in vehicle Re-Identification (ReID) in surveillance videos have focused on harnessing deep learning techniques \cite{eckstein2020large, he2021transreid}, domain adaptation strategies \cite{lee2020strdan, wang2021inter}, and multi-modal fusion approaches \cite{zhao2021heterogeneous} to enhance accuracy and robustness. These methods employ advanced feature representations, utilizing CNNs or Vision Transformers (ViTs) to capture distinctive vehicle features. Domain adaptation specifically aims to mitigate discrepancies across different camera views, while multi-modal fusion leverages combined sensor data to improve detection performance.

Evaluation of these technologies is facilitated by benchmark datasets such as VeRi \cite{liu2016deep}, VehicleID \cite{liu2016deep2}, CityFlow \cite{tang2019cityflow}, and AI City Challenge \cite{yao2020simulating}, which use metrics like rank-1 accuracy and mean Average Precision (mAP). Despite these technological strides, challenges remain, particularly concerning appearance variation and scalability across extensive network systems.

\begin{figure}[t!]
  \centering
  \includegraphics[scale=0.24]{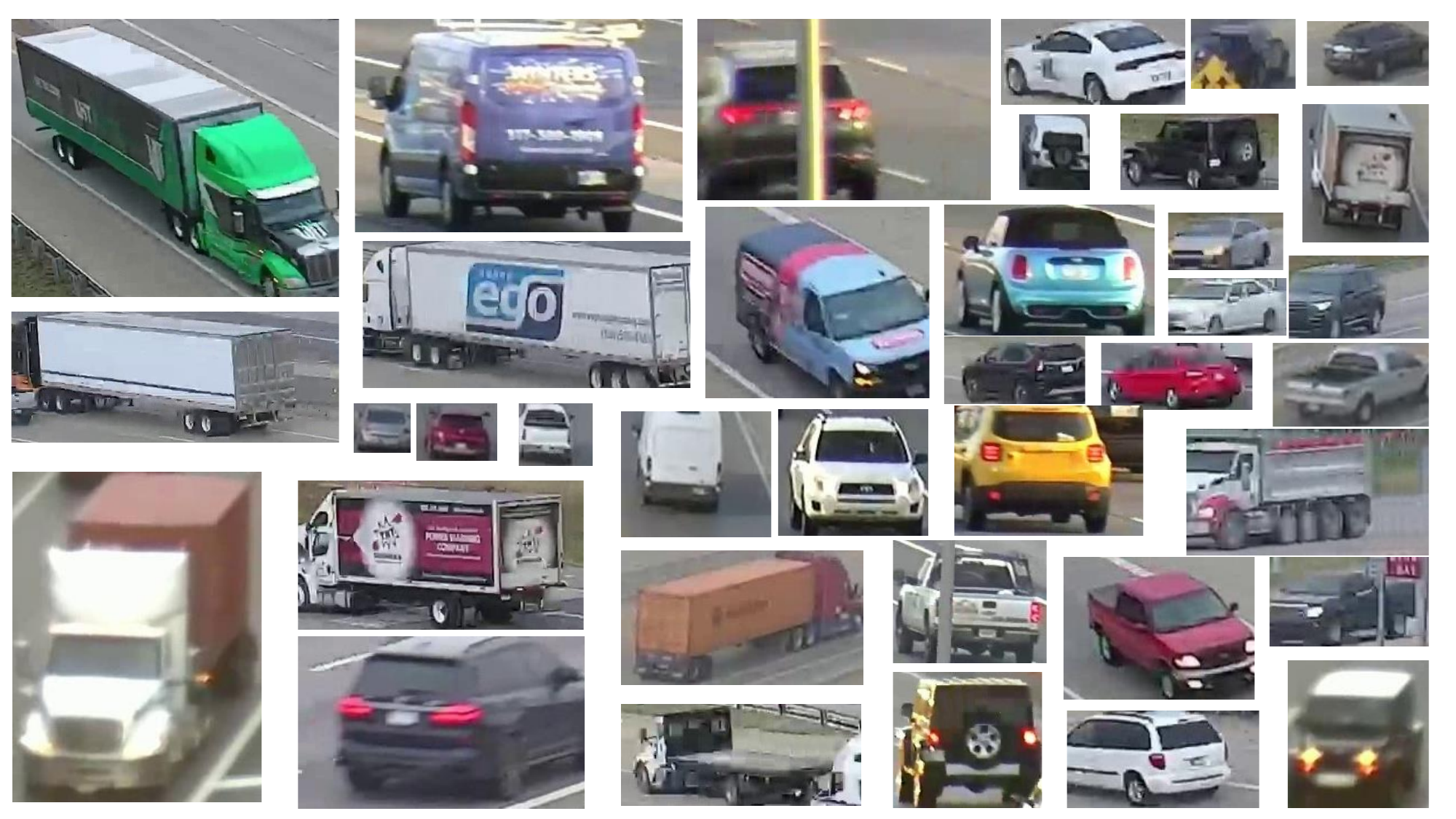}
    \caption{\textit{Some examples extracted from our Weaving ReID dataset. All the images keep their original size, shape and aspect ratio.  For each unique vehicle, it has about 15 image samples.
}}
  \label{fig:veh_example}
\end{figure}


\section{Weaving Dataset and ReID Model}\label{sec:method1}


\begin{table}[]
\fontsize{6.5}{11}\selectfont
\centering
\begin{tabular}{|c|cccc|c|}
\hline
\textbf{Dataset} &
  \textbf{\begin{tabular}[c]{@{}c@{}}VehicleID\\ \cite{liu2016deep2}\end{tabular}} &
  \textbf{\begin{tabular}[c]{@{}c@{}}VeRi-776 \\  \cite{liu2016deep}\end{tabular}} &
  \textbf{\begin{tabular}[c]{@{}c@{}}CityV2-\\ ReID\\ \cite{yao2020simulating}\end{tabular}} &
  \textbf{\begin{tabular}[c]{@{}c@{}}VERI-\\ Wild 2.0\\ \cite{bai2021disentangled}\end{tabular}} &
  {\color[HTML]{333333} \textbf{Ours}} \\ \hline
\textbf{\#Total} &
  221,763 &
  49,360 &
  85,058 &
  825,042 &
  {\color[HTML]{333333} 78,979} \\ \hline
\textbf{\#Training} &
  100,182 &
  37,778 &
  52,717 &
  277,797 &
  {\color[HTML]{333333} 69,274} \\ \hline
\textbf{\#Query} &
  - &
  1,678 &
  1,103 &
  10,000 &
  {\color[HTML]{333333} 970} \\ \hline
\textbf{\#Gallery} &
  20,038 &
  11,579 &
  31,238 &
  398,728 &
  {\color[HTML]{333333} 9,705} \\ \hline
\textbf{\begin{tabular}[c]{@{}c@{}}\%Images size \\ \textless{}200 pixels\end{tabular}} &
  0.02\% &
  35\% &
  53\% &
  \textless{}0.6\% &
  {\color[HTML]{333333} 86\%} \\ \hline
\textbf{\begin{tabular}[c]{@{}c@{}}\%Images size \\ \textgreater{}500 pixels\end{tabular}} &
  14.5\% &
  4.3\% &
  7.5\% &
  \textgreater{}30\% &
  {\color[HTML]{333333} 1.7\%} \\ \hline
\textbf{\begin{tabular}[c]{@{}c@{}}Data\\  Resources(C/D)\end{tabular}} &
  \begin{tabular}[c]{@{}c@{}}20/- \\  \end{tabular} &
  \begin{tabular}[c]{@{}c@{}}20/-\\ \end{tabular} &
  \begin{tabular}[c]{@{}c@{}}46/- \\ \end{tabular} &
  \begin{tabular}[c]{@{}c@{}}274/-\\  \end{tabular} &
  {\color[HTML]{333333} \begin{tabular}[c]{@{}c@{}}16/2\end{tabular}} \\ \hline
\end{tabular}
\caption{Comparisons Among our collected Weaving ReID Dataset and other public datasets. There are two data resources: Highway camera (C) and drone camera(D). `-' means `0'. }
\label{table:t1}
\end{table}
\smallskip
\noindent
\textbf{Weaving ReID dataset.}
Creating a specialized dataset tailored to the unique characteristics and challenges of our highway weaving areas is essential. This approach ensures that data collection specifically addresses the complexities of weaving scenarios, supporting the development of specialized analytical models. Our dataset includes images of vehicles from these areas, optimized for detection and tracking. It is enriched with manually verified details such as unique identifiers and lane IDs, as discussed in \cite{Mei2024}. Three graduate students invested over 120 hours each in labeling and cleaning the data, employing a majority voting strategy informed by the travel time formula $\Delta t = \frac{S}{V}$, where $S$ represents the average distance and $V$ represents the average vehicle speed between the entry and exit points of the weaving areas.

Our comprehensive dataset features 4,902 unique vehicles across nine weaving areas, captured using both highway and drone cameras. With its varied image sizes and resolutions, and mixed sources of imaging, this dataset presents more significant challenges than standard datasets. It is systematically divided into training, query, and gallery subsets, as detailed in Table \ref{table:t1} and illustrated in Figure \ref{fig:veh_example}, offering a valuable resource for both research and practical applications.

\smallskip
\noindent
\textbf{Weaving ReID Model.} Our weaving ReID model is designed to recognize the same vehicle across different weaving areas, despite variations in lighting, angles, or partial obstructions. This capability is vital for analyzing traffic flow and understanding complex driving behaviors, such as weaving through traffic. The model learns from our labeled dataset of vehicle images, extracting and utilizing robust, distinctive features that enable accurate vehicle matching across varying camera inputs and conditions.

Our vehicle Re-ID framework utilizes a pure Vision Transformer (ViT) with the pre-trained backbone ``vit\_base\_patch16\_224". We adopt the "Transformer-based strong baseline framework" from TransReID \cite{he2021transreid} as our structural baseline. Feature extraction leverages the pre-trained weights, while we fine-tune it on our weaving ReID dataset. This approach helps leverage the generalization capabilities of large-scale pre-trained models, enhancing performance on the specialized task of re-identification by adapting vision transformer to the nuances and specific features required for our ReID challenges.


For the patch partition, we maintain the patch size of $p = 16$, consistent with the pre-trained ViT-B/16, and use a even stride ($s = 16$) in each dimension. 
Given an image with dimensions Height, Width, Color Channel: $H \times W \times CP$ , it is then cut into patches of size $p \times p$ with a non-overlap defined by the an even stride $s$, then we will get $n$ patches, and \( n \) is calculated as $\left(\frac{H}{s} \right) \times \left(\frac{W}{s}\right)$. 
During our fine-tuning $L_{id}$ and $L_{triplet}$ losses are used to optimize the weights in the ViT network:
\begin{equation}
    L = \lambda_{id} L_{id} + \lambda_{triplet} L_{triplet} 
\end{equation}
and 
\begin{equation}
L_{id}= -\sum_{i=1}^{N} \sum_{c=1}^{C} y_{i,c} \log(\hat{y}_{i,c})
\end{equation}
where $C$ is the number of classes, 
$y_{i,c}$ is the ground truth probability (1 for the true class and 0 for others), and 
$\hat{y}_{i,c}$ is the predicted probability for class $c$ for the i-th example.
\begin{equation}
    L_{triplet}= \frac{1}{N} \sum_{i=1}^{N} \log \Big[ 1 + \sum_{j=1}^{N} \exp \left( d(a_i, p_i) - d(a_i, n_j) \right) \Big]
\end{equation}
where $L_{triplet}$ is the soft triplet loss over the batch. $N$ is the number of samples in each batch, $d(a,b)$ is the distance between the embeddings of 
$a$ and 
$b$, which is often computed as a  Euclidean distance for the embedding space.

\section{Weaving Analysis}\label{sec:method2}

\smallskip
\noindent
\textbf{ReID Feature Extraction and Matching.}
Vehicle matching occurs using cosine similarity between features extracted from our trained ReID network.
Let $\mathbf{v}_1$ and $\mathbf{v}_2$ be two feature vectors of two images extracted from our ReID model.
The cosine similarity between $\mathbf{v}_1$ and $\mathbf{v}_2$ is given by:
\begin{equation}
\text{cosine\_similarity}(\mathbf{v}_1, \mathbf{v}_2) = \frac{\mathbf{v}_1 \cdot \mathbf{v}_2}{\|\mathbf{v}_1\| \|\mathbf{v}_2\|}
\end{equation}

where $\mathbf{v}_1 \cdot \mathbf{v}_2$ denotes the dot product of $\mathbf{v}_1$ and $\mathbf{v}_2$, and $\|\mathbf{v}\|$ denotes the Euclidean norm of vector $\mathbf{v}$. In vehicle re-identification, the larger the value of cosine similarity, the higher the probability that the two vehicles are the same.

\smallskip
\noindent
\textbf{Hungarian Matching.} After conducting deep feature matching, we observed that almost all matched vehicle pairs had similar travel times to the exit. To reduce false positives while retaining true positive matches after the ReID feature matching, we have implemented a specialized cost matrix within the standard Hungarian Matching algorithm. This algorithm uses a uniquely crafted cost matrix $M$, formulated to assess all possible vehicle pairings ($v_i$ and $v_j$) from origin-destination weaving areas with average driving time $T_a$.
The cost matrix $M$ can be constructed as follows $M[i,j]$:
\begin{equation}
\begin{cases} 
  w_1 \cdot d(f(v_{1i}), f(v_{2j})) + \\w_2 \cdot |t(v_{1i}) - t(v_{2j}) - T_a|  & \text{if} 
  \begin{aligned}[t]
&r(v_{1i}, v_{2j}, c_k) = 1\\
\text{and}\\
&t(v_{1i})-t(v_{2j})<0
\end{aligned}\\
\infty & \text{otherwise}
\end{cases}
\end{equation}

In this formulation:
\begin{itemize}
  \item $M[i, j]$ represents the cost of matching vehicle $v_{1i}$ from the first area with vehicle $v_{2j}$ from the second area.
  \item $d(f(v_{1i}), f(v_{2j}))$ denotes the cosine distance between features of $v_{1i}$ and $v_{2j}$ extracted from the ReID network.
  \item The function $r(v_{1i}, v_{2j}, c_k)$ serves as a filter to ensure that vehicles detected by two different cameras, and identified as either a ``car" or a ``truck", are indeed the same vehicle type $c_k$.
  \item If matching vehicle $v_{1i}$ with $v_{2j}$ to class $c_k$ is compatible based on other filters ($r(v_{1i}, v_{2j}, c_k) = 1$), and $t(v_{1i}) - t(v_{2j}) < 0 $, the cost is the combination of the distance between the feature vectors of $v_{1i}$ and $v_{2j}$, the absolute difference of their travel times and the average travel time $T_a$, weighted by $w_1$ and $w_2$ respectively.
  \item If matching $v_{1i}$ with $v_{2j}$ to class $c_k$ is not compatible or $t(v_{2j})$ is less than $t(v_{1i})$, the cost is set to infinity, indicating that the assignment is not feasible due to the timing.
\end{itemize}

\begin{figure}[t!]
  \centering
  \includegraphics[scale=0.27]{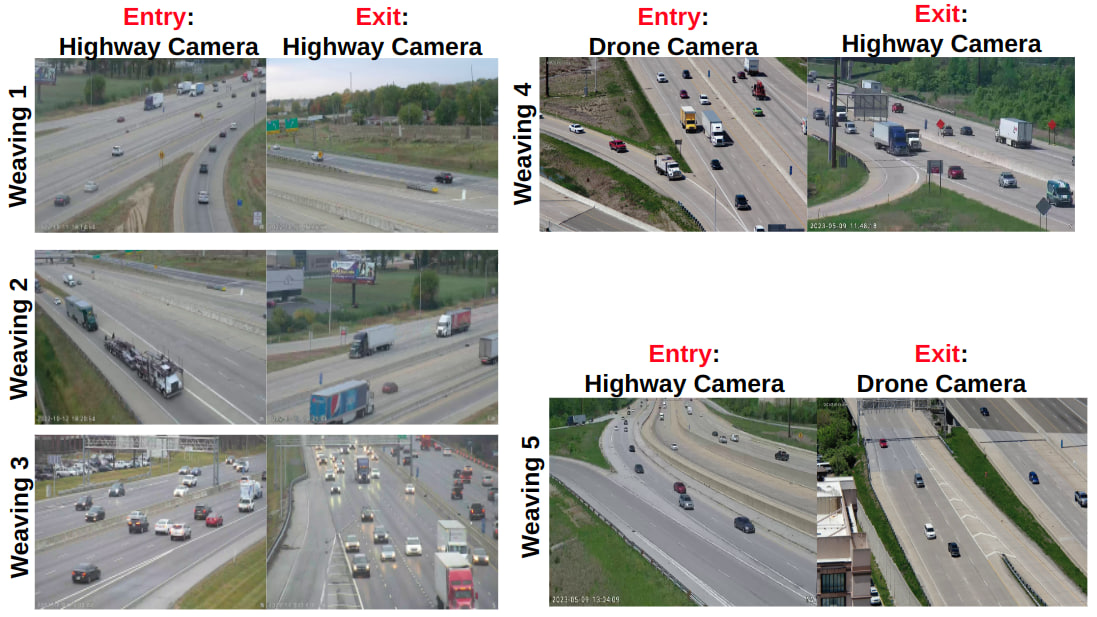}
    \caption{\textit{Videos from five weaving areas are used in our experiments. 
}}
  \label{fig:weaving_example}
\end{figure}

\smallskip
\noindent
\textbf{Lane-wise Traffic Pattern Estimation.}
Due to technological limitations, only a subset of vehicles traveling from P1 to P2 can be successfully matched. However, we can calculate the ratio of matched vehicles to the total number of vehicles passing through each pair of entry and exit lanes. These ratios function as statistical sampling results. By accurately counting the total number of vehicles passing through each lane at both entry (P1) and exit (P2) points, and considering the set of matched vehicle pairs as representative samples, we can use these results to estimate lane-specific traffic patterns across each weaving area.

\section{Experiments}
\subsection{Experiment settings}
\smallskip
\noindent
\textbf{Dataset.}
We captured videos from all nine weaving areas to create the ground truth for vehicle matching, forming the core of our focused weaving ReID dataset—a subset of the complete ground truth. To assess the accuracy of our method, we specifically analyzed data from five weaving areas, as depicted in Figure \ref{fig:weaving_example}. For weaving areas 1, 2, and 5, both entry and exit surveillance was conducted using highway cameras. In contrast, weaving area 3 used a drone camera for entry monitoring paired with a highway camera for exit surveillance, while weaving area 4 utilized a highway camera at the entry point and a drone camera at the exit. All footage was captured during daylight hours—specifically in the morning, at noon, and in the late afternoon, with video resolutions ranging from 352x240 to 1980x1080, and each session recorded over the last 10 minutes.

\smallskip
\noindent
\textbf{Evaluation Metrics for Weaving Analysis.}
We use True Positive Rate (TPR) and Precision to describe the quality of the weaving analysis results.

  
\begin{figure}[t!]
  \centering
  \includegraphics[scale=0.5]{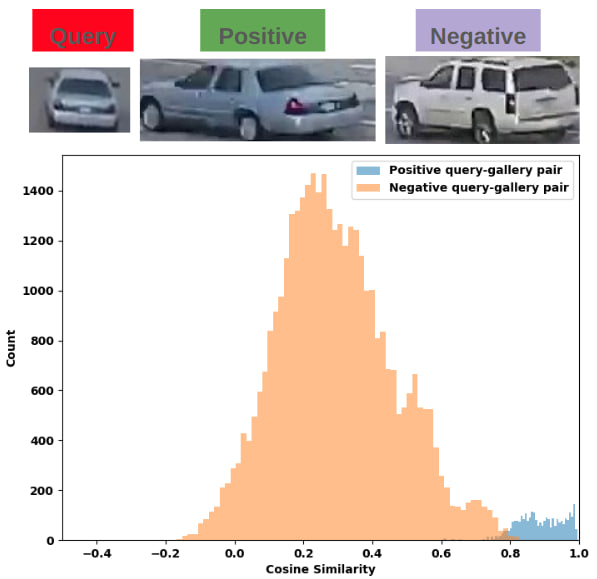}
    \caption{\textit{Feature Similarity Analysis involves extracting feature vectors for query, positive, and negative vehicle samples from our ReID model. The cosine similarity between positive pairs is significantly higher than that between negative pairs, providing a robust criterion for vehicle matching.
}}
  \label{fig:cossine}
\end{figure}
\begin{figure*}[t!]
  \centering
  \includegraphics[scale=0.5]{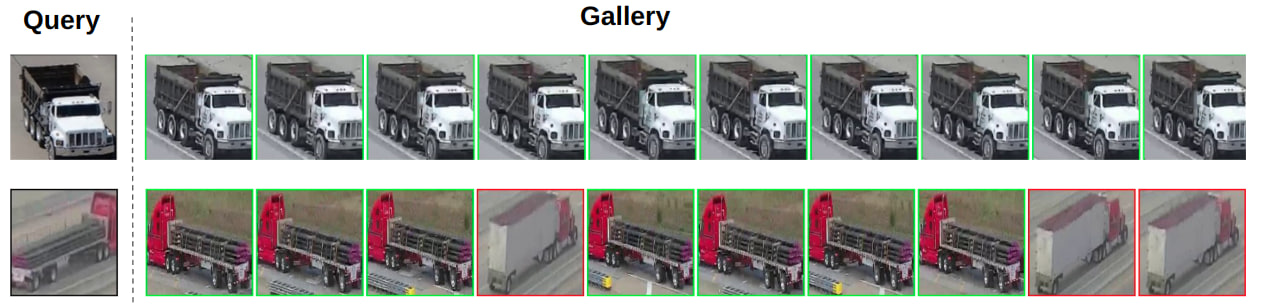}
  \caption{\textit{For each query vehicle sample, the top 10 matching results are presented with green outlines indicating correct matches and red outlines highlighting incorrect matches}.}
  \label{fig:tk10}
\end{figure*}
\begin{figure}[t!]
  \centering
  \includegraphics[scale=0.6]{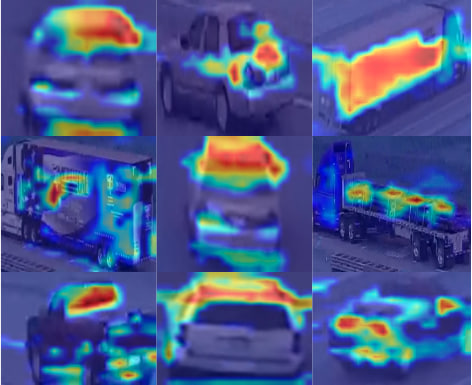}
  \caption{\textit{Grad-CAM visualization of attention maps. Our model can learn both global and local information well}.}
  \label{fig:cam}
\end{figure}

\smallskip
\noindent
\textbf{Implementation Details.}
Training and testing were conducted on an NVIDIA RTX A6000 GPU utilizing PyTorch toolbox 1 for FP16 training. We initialized the ``vit\_base\_patch16\_224" model with ImageNet pre-trained parameters, adapting the final layer to $N$ dimensions to match the number of identities in our dataset. Training batches consisted of $4$ identities with $16$ images per identity (batch size $B = 64$), resized to $224 \times 314$ pixels, zero-padded, and cropped. Patch size was set to 16 with strides of $[12, 16]$. Both $L_{id}$ and $L_{triplet}$ were set to 1.
During testing, images retained their original size, with a 50\% probability of horizontal flipping, and were normalized. Training employed an SGD optimizer with a momentum of 0.1 and weight decay of 1e-4, spanning 120 epochs with an initial learning rate of 0.035. Feature extraction took place before the BN(batch normalization) layer in inference.

In the matching phase, mismatches were filtered using a cosine similarity threshold of 0.8. For Hungarian Matching, weights $W_1$ and $W_2$ were set to 0.3 and 0.75, respectively.

\subsection{Results}

\smallskip
\noindent
\textbf{Vehicle Re-identification and Feature Matching.}
In our testing subset, we achieved a mean Average Precision (mAP) of 47.8\%, with Cumulative Matching Characteristic (CMC) scores of 42\% for Rank 1, 50.9\% for Rank 5, and 57.2\% for Rank 10. Several matching samples are depicted in Figure \ref{fig:tk10}.
As depicted in Figure \ref{fig:cossine}, the positive pairs of vehicles' features extracted from our weaving ReID model demonstrate significantly higher similarity compared to those from negative pairs. This observation underscores our model's exceptional capability to discern distinct vehicle features from our dataset. We utilize the Grad-CAM visualization method \cite{selvaraju2017grad} to highlight the specific regions of input vehicles that our model focuses on, as depicted in Figure \ref{fig:cam}.

\smallskip
\noindent
\textbf{Weaving Analysis.}
Table \ref{table:t2} demonstrates generally satisfactory counting accuracy, although weaving 2 exhibits lower performance, potentially attributed to a static IOU threshold for tracking. Implementing an adaptive threshold could potentially enhance performance. Highway cameras, typically mounted on trusses or poles, offer limited angles, often capturing different vehicle sides at entry/exit points. In 7 out of 8 cases, one camera captures the vehicle's front while the other captures the rear, contributing to varied precision, particularly evident in weaving 3. Combining front and side views could enhance identification accuracy. Utilizing a combination of fixed and drone cameras, or employing dual drones, could optimize viewing angles and improve overall performance. Notably, drone footage from weaving 4 and 5 exhibits superior quality and precision, underscoring the significant impact of camera quality on results.

\begin{table}[]
\centering
\resizebox{\columnwidth}{!}{%
\begin{tabular}{|c|c|c|c|c|c|}
\hline
\textbf{Weaving Area} &
  \textbf{\begin{tabular}[c]{@{}l@{}}Time of \\ the Day\end{tabular}} &
  \textbf{\begin{tabular}[c]{@{}l@{}}Vehicle visible\\ Side\end{tabular}} &
  \textbf{\begin{tabular}[c]{@{}l@{}}Vehicle Count\\ Accuracy\%\end{tabular}} &
  \textbf{TPR\%} &
  \textbf{\begin{tabular}[c]{@{}l@{}}Precision\\ \%\end{tabular}} \\ \hline
\textbf{1} & M & RS-RS & 93  & 34    & 84.3  \\ \hline
\textbf{1} & N & RS-RS & 100 & 45.29 & 94.7  \\ \hline
\textbf{1} & A & FS-FS & 98  & 33.97 & 74.64 \\ \hline
\textbf{2} & M & FS-FS & 82  & 42.01 & 77.17 \\ \hline
\textbf{2} & A & FS-FS & 78  & 40.69 & 69.53 \\ \hline
\textbf{3} & N & RS-F  & 95  & 22.4  & 53.16 \\ \hline
\textbf{4} & N & FS-FS & 99  & 27.65 & 71.85 \\ \hline
\textbf{5} & N & FS-FS & 100 & 35.28 & 88.46 \\ \hline
\end{tabular}%
}
\caption{Accuracy of weaving analysis. M = Morning, N = Noon, A = late Afternoon, F = front, R = rear, S = side, FS = front and side, RS = rear and side, TPR is the percentage of system-identified vehicle matches in all vehicles detected, Precision is the percentage of correct matches in all system-identified matches.}
\label{table:t2}
\end{table}


\section{Conclusion}
Lane-based highway traffic weaving analysis is vital for understanding vehicle lane-change patterns, aiding traffic management and road design. This paper treats weaving analysis as a vehicle-matching sampling issue. Our experiments show that features from our ReID model are highly distinguishable, enhancing vehicle matching accuracy and advancing weaving analysis. Future work will refine our ReID model for improved performance.

\smallskip
\noindent
\textbf{Acknowledgment}.
This work was supported by the Joint Transportation Research Program (JTRP), administered by the Indiana Department of Transportation and Purdue University, Grant SPR-4738.

{\small
\bibliography{egbib}
}




\end{document}